\documentclass[conference]{IEEEtran}
\ifCLASSINFOpdf
   \usepackage[pdftex]{graphicx}
\else
\fi
\ifCLASSOPTIONcompsoc
  \usepackage[caption=false,font=normalsize,labelfont=sf,textfont=sf]{subfig}
\else
  \usepackage[caption=false,font=footnotesize]{subfig}
\fi
\usepackage[hidelinks]{hyperref}

\usepackage[acronym]{glossaries}
\glsdisablehyper %Disable hyperlink
\usepackage{caption}

\newacronym{cyclegan}{CycleGAN}{Cycle-consistent Generative Adversarial Network}
\newacronym{gan}{GAN}{Generative Adversarial Network}
\newacronym{cycada}{CyCADA}{Cycle-Consistent Adversarial Domain Adaptation}
\newacronym{fid}{FID}{Frechet Inception Distance}
\newacronym{kid}{KID}{Kernel Inception Distance}
\newacronym{hgan}{H-GAN}{HandGAN}
\newacronym{geocongan}{GeoConGAN}{Geometrically consistent CycleGAN}
\newacronym{vit}{ViT}{Vision Transformer}

\usepackage{array}
\usepackage{booktabs}
\usepackage{amssymb}
\usepackage{pifont}
\usepackage[para,online,flushleft]{threeparttable}
\usepackage{multirow}

\newcolumntype{M}[1]{>{\centering\arraybackslash}m{#1}}
\newcommand{\ra}[1]{\renewcommand{\arraystretch}{#1}}

\newcommand{\comment}[1]{}

\usepackage{tikz}
\usetikzlibrary{calc}

\usepackage{nicefrac,xfrac}
\usepackage{textcomp}

\begin{document}
\title{H-GAN: the power of GANs in your Hands}

%\author{\IEEEauthorblockN{Sergiu Oprea}
%\IEEEauthorblockA{School of Electrical and\\Computer Engineering\\
%Georgia Institute of Technology\\
%Atlanta, Georgia 30332--0250\\
%Email: http://www.michaelshell.org/contact.html}
%\and
%\IEEEauthorblockN{Homer Simpson}
%\IEEEauthorblockA{Twentieth Century Fox\\
%\and
%\IEEEauthorblockN{James Kirk\\ and Montgomery Scott}
%\IEEEauthorblockA{Starfleet Academy\\
%San Francisco, California 96678--2391\\


\author{\IEEEauthorblockN{Sergiu Oprea\IEEEauthorrefmark{1},
Giorgos Karvounas\IEEEauthorrefmark{2},
Pablo Martinez-Gonzalez\IEEEauthorrefmark{1},
Nikolaos Kyriazis\IEEEauthorrefmark{2},
Sergio Orts-Escolano\IEEEauthorrefmark{3},\\
Iason Oikonomidis\IEEEauthorrefmark{2},
Alberto Garcia-Garcia\IEEEauthorrefmark{4},
Aggeliki Tsoli\IEEEauthorrefmark{2},
%John Alejandro Castro-Vargas\IEEEauthorrefmark{1},\\
Jose Garcia-Rodriguez\IEEEauthorrefmark{1}, and
Antonis Argyros\IEEEauthorrefmark{2}}
\IEEEauthorblockA{\IEEEauthorrefmark{1} Department of Computer Technology, University of Alicante, Spain \\
Email: \{soprea, pmartinez, jgarcia\}@dtic.ua.es}
\IEEEauthorblockA{\IEEEauthorrefmark{2}Institute of Computer Science, FORTH, Greece \\
Email: \{gkarv, kyriazis, oikonom, aggeliki, argyros\}@ics.forth.gr}
\IEEEauthorblockA{\IEEEauthorrefmark{3} Department of Computer Science and Artificial Intelligence, University of Alicante, Spain \\
Email: sorts@ua.es}
\IEEEauthorblockA{\IEEEauthorrefmark{4} Institute of Space Sciences (ICE-CSIC), Spain \\
Email: garciagarcia@ice.csic.es}}


\twocolumn[{%
\renewcommand\twocolumn[1][]{#1}%
\maketitle

\begin{center}
    \centering
    \includegraphics[width=0.95\textwidth]{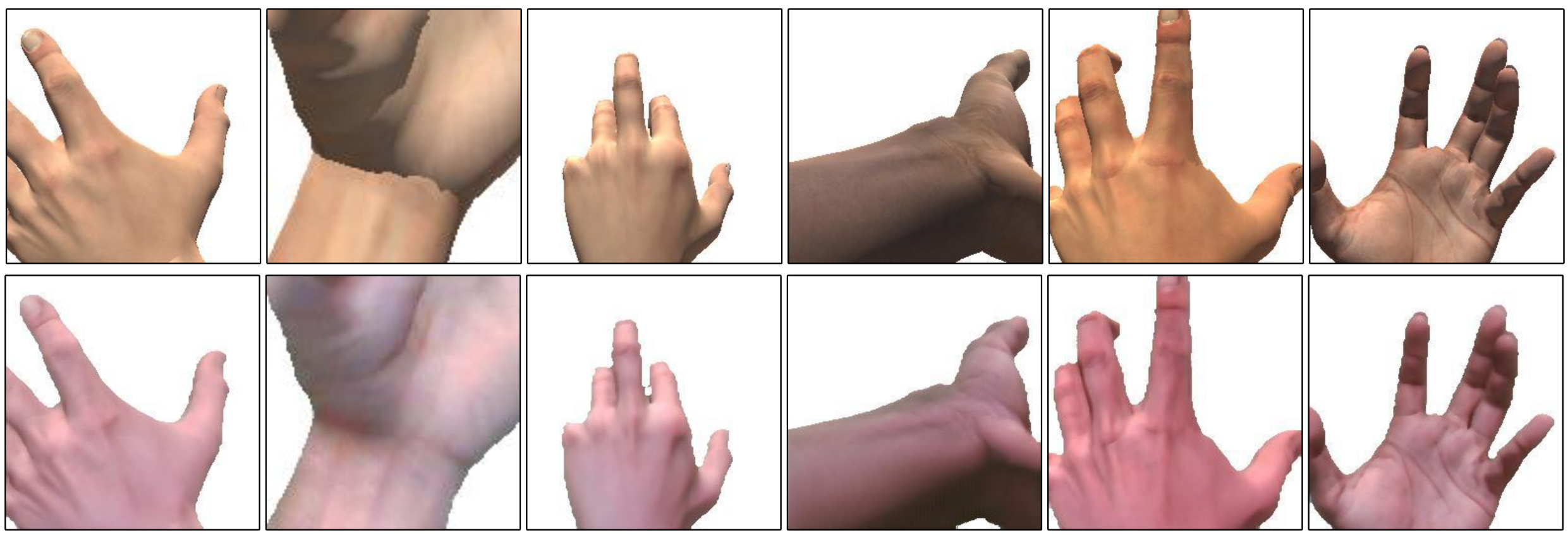}
    \captionof{figure}{Top row, images from the SynthHands dataset~\cite{Mueller2017}. Bottom row, synthetic hands translated to the real domain preserving their shape. Images were generated using our msH-GAN4 model.}
    \label{fig:hands_teaser}
\end{center}
}]

%To this end, hand shape is preserved in the translation process to ensure inter-domain annotation correspondence.
\begin{abstract}
We present \gls{hgan}, a cycle-consistent adversarial learning approach implementing multi-scale perceptual discriminators. It is designed to translate synthetic images of hands to the real domain. Synthetic hands provide complete ground-truth annotations, yet they are not representative of the target distribution of real-world data. We strive to provide the perfect blend of a realistic hand appearance with synthetic annotations. Relying on image-to-image translation, we improve the appearance of synthetic hands to approximate the statistical distribution underlying a collection of real images of hands. \mbox{\gls{hgan} tackles} not only the cross-domain tone mapping but also structural differences in localized areas such as shading discontinuities. Results are evaluated on a qualitative and quantitative basis improving previous works. Furthermore, we relied on the hand classification task to claim our generated hands are statistically similar to the real domain of hands.
\end{abstract}

\begin{IEEEkeywords}
hands, synthetic-to-real, generative adversarial networks, cycle-consistency, perceptual discriminator, multi-scale, vision transformer
\end{IEEEkeywords}

%For this purpose, our methods tackles not only skin color mapping but also structural differences in localized areas. We show improvements from tone mapping to  


% \fi
\IEEEpeerreviewmaketitle

\section{Introduction}
%Contextualization of the problem to be addressed
The lack of large amounts of high-quality annotated data is still a barrier for supervised deep learning approaches. Manual labelling is tedious and time-consuming, particularly when it comes to per-pixel or 3D annotations. For this reason, generating fully annotated and quasi-unlimited data from a controlled virtual environment emerged as a potential solution to data scarcity. Synthetic data generation, despite its recent success in different fields~\cite{Nikolaus2018,Garcia2018a,Hwang2020}, has at least two major flaws: i) it is challenging, if not impossible, to perfectly simulate real world properties in a synthetic environment, and ii) training deep models exclusively on synthetic data (even if they are photorealistic) commonly exhibits generalization issues in the real domain. In the context of domain adaptation~\cite{Zhao2020}, these issues are addressed by reducing the domain distribution discrepancy between synthetic and real data. 

\vspace*{0.1cm}\noindent\textbf{Purpose:} We focus on changing the visual appearance of synthetically generated hands to approximate the statistical distribution underlying a collection of real images of hands. We focus on hands for two reasons: i) synthetic hands provide full ground-truth annotations, yet they are not representative of the target distribution of real-world data, and ii) large-scale datasets of real hands are sparsely labeled e.g. they lack 3D annotations. In this work, we want to translate synthetic hands to the real domain, preserving the hand shape to ensure the inter-domain ground truth correspondence (see Figure~\ref{fig:hands_teaser}).

\vspace*{0.1cm}\noindent\textbf{Our work:} For this purpose, we present the \gls{hgan} architecture, a cycle-consistent adversarial learning approach~\cite{Zhu2017} implementing multi-scale perceptual discriminators~\cite{Sungatullina2018}. On the one hand, cycle-consistency makes the model learn the mapping between the two domains in an unsupervised fashion and from unpaired images. On the other hand, perceptual discriminators enforce cross-domain content and style transfer by learning statistics in a high-level feature space. Besides tone mapping, this synergy allows the model to hallucinate finer details (see Figure~\ref{fig:hair}) and remove artifacts such as shading discontinuities (see Figure~\ref{fig:artifacts}). This leads to an overall improvement in the appearance of the generated hands, which we assess on a qualitative and quantitative basis. Our model is closely related to~\cite{Mueller2018}, sharing the same goal as their \gls{geocongan} model.

\vspace*{0.1cm}\noindent\textbf{Contributions:} First, we present an adversarial approach for unsupervised domain adaptation reducing the domain gap discrepancy between synthetic and real hands. Second, we show improvements over the proposed baseline~\cite{Mueller2018} on a qualitative and quantitative basis. Furthermore, relying on the hand classification task we claim our generated hands are statistically similar to the real domain of hands. Our source code is available at GitHub\footnote{\href{https://github.com/sergiuoprea/hgan}{https://github.com/sergiuoprea/hgan}}.

\section{Related Works}
\label{sec:related_works}
In this section we provide a brief overview of works on domain adaptation, perceptual losses and synthetic-to-real translation as they are related topics to our task at hand.

\vspace*{0.1cm}\noindent\textbf{Domain adaptation} aims to narrow the distribution discrepancy between a labeled source domain and an unlabeled or sparsely annotated target domain~\cite{Zhao2020}. Hoffman \textit{et al.} proposed the \gls{cycada} architecture~\cite{Hoffman2018}. To preserve semantic information in the distribution alignment process, they used a semantic task loss to enforce the cycle-consistency. Likewise, Tran \textit{et al.}~\cite{Tran2019} proposed an attribute-conditioned \gls{cyclegan} generating multiple target images with different attributes e.g. day or night. Aiming to translate domain-invariant features, Li \textit{et al.}~\cite{Li2019} proposed a more computationally efficient approach than~\cite{Hoffman2018} reporting also better results. Baek \textit{et al.}~\cite{baek2020weakly} proposed a framework based on \glspl{gan} and mesh rendering, that adapts the hand-object domain to a hand-only domain, while it learns hand pose estimation. Rad \textit{et al.}~\cite{rad2018domain} using depth information, showed accurate hand pose estimation on RGB images. For a deeper insight, we recommend the reader to consider~\cite{Zhao2020} reviewing the recent progress on single-source unsupervised domain adaptation.

\vspace*{0.1cm}\noindent\textbf{Synthetic-to-real translation} is the task of domain adaptation from synthetic data to real data. Advances in this field have enabled the use of quasi-unlimited and fully-labelled synthetically generated data in real-world tasks. Such tasks include vehicle re-identification~\cite{Lee2020}, depth estimation~\cite{Maximov2020}, and geometry estimation~\cite{PNVR2020}, among others. Closely related to our work, Mueller \textit{et al.}~\cite{Mueller2018} used a geometrically-consistent \gls{cyclegan} for the same goal of closing the gap between synthetic and real images of hands. Concretely, they have used the visually improved and fully-annotated synthetic hands for the 3D hand tracking task.  Shrivastava \textit{et al.}~\cite{Shrivastava2017} proposed a refiner network to improve synthetically generated images of eyes. Likewise, Atapattu \textit{et al.}~\cite{Atapattu2019} addressed the same task. On the other side, deep image synthesis showed relevant progress for the task at hand. Sangkloy \textit{et al.}~\cite{Sangkloy2017} propose the Scribbler method, an adversarial autoencoder design to synthesize images from imperfect hand drawn sketches using sparse color scribbles. Chen and Koltun~\cite{Chen2017} showed in their work that given a semantic layout, their proposed model is able to synthesize photographic images. The network is trained in a supervised fashion on pairs of photographs and their corresponding semantic layouts.

\vspace*{0.1cm}\noindent\textbf{Perceptual losses} use activations of pretrained networks such as VGG~\cite{Simonyan2015} to train another network. Such features, extracted at different scales, represent not only low-level image details, but also global structure. This has fostered advances in image style transfer and super-resolution~\cite{Johnson2016,Wang2020}, texture synthesis~\cite{Ulyanov2016}, image inpainting~\cite{Su2019} and image autoencoder embeddings~\cite{Pihlgren2020}, to name a few. Perceptual losses have been widely used in supervised tasks, however in the unsupervised scenario their application is not direct because of the lack of input/target image pairs. In this regard, Wang \textit{et al.}~\cite{Wang2018} was the first to implement a multi-scale perceptual discriminator for the unsupervised image-to-image translation task. PerceptualGAN~\cite{Li2017} uses perceptual losses to narrow the representation difference between small and large-sized objects to improve small object detection. Correlating perceptual similarity between pairs of images with human perceptual judgement, Zhang \textit{et al.}~\cite{Zhang2018} found that activations extracted from intermediate layers of a classification network do indeed correspond to human perception.

\section{Method}
\begin{figure*}[!htb]
\centering
\includegraphics[width=\textwidth]{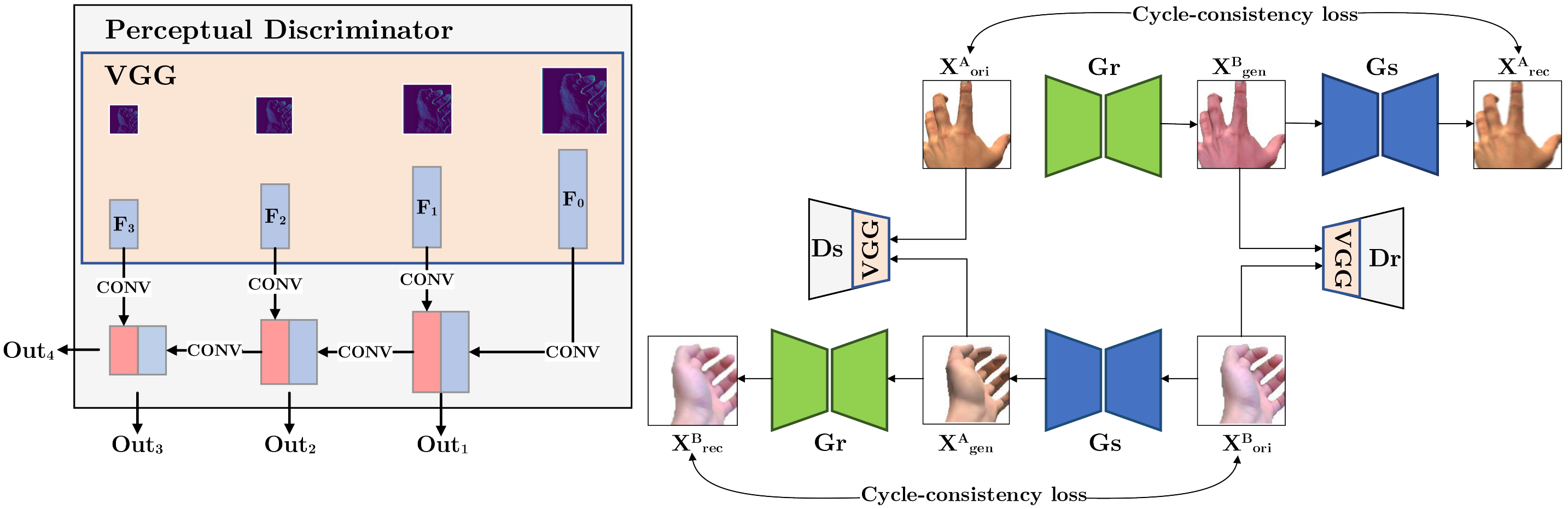}
\caption{On the left, multi-scale perceptual discriminator using features, $F_i$, extracted at different scales from VGG network. At each scale, perceptual statistics are stacked together after convolutional blocks (convolution + leakyReLU). On the right, our \gls{hgan} model performing the inter-domain translation in both directions. On the one side, generators $G_r$ and $G_s$ perform the synthetic-to-real and real-to-synthetic translation, respectively. On the other side, discriminators learn the statistics underlying real ($D_r$) and synthetic ($D_s$) domains. The generators and discriminators are trained simultaneously, in an adversarial fashion. The representation of the perceptual discriminator was inspired by~\cite{Sungatullina2018}.}
\label{fig:architecture}
\end{figure*}

\begin{figure}[!htb]
\centering
\includegraphics[width=\linewidth]{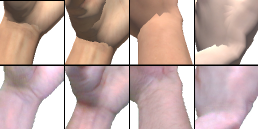}
\caption{Top row, synthetic hands. Bottom row, translated hands to the real domain. Notice how the perceptual discriminator drives the generator to treat shading discontinuities at the wrist of the hand, which are not found in real data.}
\label{fig:artifacts}
\end{figure}

\begin{figure}[!htb]
\centering
\includegraphics[width=\linewidth]{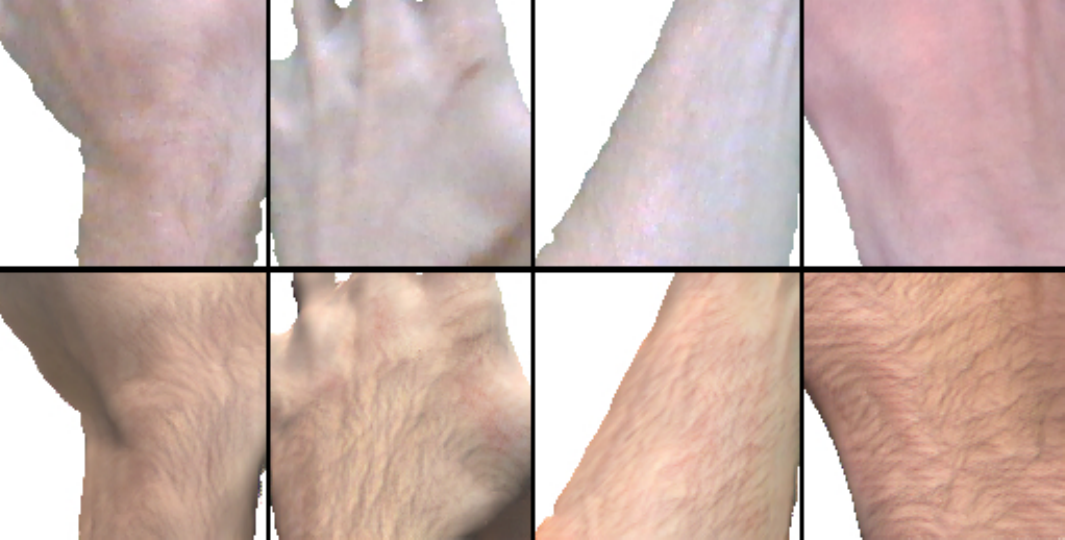}
\caption{Top row, real hands. Bottom row, hands translated to the synthetic domain. Notice how the perceptual discriminator makes the generator to hallucinate hair in the synthetic domain.}
\label{fig:hair}
\end{figure}

We introduce \acrfull{hgan}, an image-to-image translation model dedicated to narrow the distribution discrepancy between synthetic and real images of hands. With this architecture, we opt to achieve the following objectives:
\begin{itemize}
    \item \emph{Learning a mapping between synthetic and real hands.} For this purpose, we rely on \glspl{cyclegan}~\cite{Zhu2017}. These architectures were specifically designed for the unsupervised image translation task given unpaired training data. Their main goal is the cross-domain transfer of low-level appearance such as color or texture while preserving high-level attributes such as content or geometric structure. Our model was built on \glspl{cyclegan} for two reasons: i)~we deal with unpaired data, since for a given image of a synthetic hand we lack its corresponding target image in the real domain and vice versa, and ii)~they excel at learning a consistent texture mapping between domains, a fact that helps preserve skin color in the translation. As a drawback, they also one-to-one map unrealistic artifacts from the synthetic domain.
    
    \item \emph{Improving the cross-domain transfer of higher-level details.}
    Synthetic hands sometimes present unrealistic lighting and shadows due to imprecise geometry, among other undesirable artifacts. Avoiding their transfer to the target domain is not straightforward. This behavior is reinforced by the per-pixel losses implemented in the vanilla~\gls{cyclegan} discriminators. To mitigate this side-effect, perceptual losses~\cite{Johnson2016} showed great potential guiding the learning process in the high-level feature space. This is, matching feature representations extracted from pretrained networks. In contrast to per-pixel losses, these features are robust and invariant to slight image deformations. Relying on high-level features, perceptual discriminators~\cite{Sungatullina2018} enforce a more precise cross-domain transfer of high-level attributes. Our model implements a perceptual discriminator striving for a consistent inter-domain mapping of high-level cues. This will reduce to some extent shading discontinuities (see Figure~\ref{fig:artifacts}) and makes the model hallucinate content from the source domain such as hair (see Figure~\ref{fig:hair}).
\end{itemize}

\subsection{\acrlong{hgan} Model}
\gls{hgan} is built on \glspl{cyclegan}~\cite{Zhu2017}, and implements multi-scale perceptual discriminators~\cite{Sungatullina2018} to enhance the translation of high-level appearance cues. Our model improves results compared to a state-of-the-art work \cite{Mueller2017}. Its architecture is depicted in Figure~\ref{fig:architecture}.

\vspace*{0.1cm}\noindent\textbf{Using \glspl{cyclegan}}: These models were designed for the unsupervised image-to-image translation task with unpaired data. They consist of two generators and discriminators (one generator and discriminator per domain) trained in an adversarial fashion. While generators perform the inter-domain translation, discriminators supervise their learning process based on the internal representation learned from each domain. The goal is to make generated data follow the underlying distribution of the target domain so as to fake discriminators.

\vspace*{0.1cm}\noindent\textbf{Inspiration by \gls{geocongan}}: On top of \gls{cyclegan} architecture, \gls{geocongan} implements a geometrically-consistent loss designed to preserve the shape of the hand in the translation. This is the binary cross-entropy between the hand shape in the source and target domains. For this purpose, it uses a pretrained network (SilNet) to segment the shape of a hand. This prevents the implementation of \gls{geocongan} as an end-to-end network, leading to additional memory overhead. The advantages of the SilNet are clear when datasets do not provide hand segmentation masks. However, its usefulness is an open question as \glspl{cyclegan} inherently preserve structure during the translation using the cycle and identity losses. If hand segmentation masks are provided, these could be used in the loss function or to mask generator's outputs as we do in our models.

\vspace*{0.1cm}\noindent\textbf{Multi-scale perceptual discriminator}: Our approach is to implement a multi-scale perceptual discriminator based on feature representations extracted at different scales from a pretrained VGG16 net~\cite{Simonyan2015}. Following the original implementation~\cite{Sungatullina2018}, it consists of stacking learnable convolutional blocks used to process activations extracted at different scales from the VGG-16. Multi-scale representations help the model to learn from data represented at different levels of abstraction, enforcing the inter-domain mapping of both low-level and high-level features.

\subsection{Implementation}

\vspace*{0.1cm}\noindent\textbf{Building blocks:} We have used the building blocks implemented in the \gls{cyclegan} architecture. \gls{cyclegan} implements PatchGAN-like~\cite{Isola2017} discriminators designed to map the input images to square patches. These discriminators learn the statistics at a patch-level, leveraging their effective receptive field. They aim to classify whether overlapping image patches come from the real or synthetic domain. As an alternative, PixelGAN ($1 \times 1$ PatchGAN) discriminators perform a per-pixel decision and in some tasks showed an enhanced diversity in generated data. On the generator side, the predominant models are ResNet-based~\cite{He2016}. We have considered both patch and pixel-based discriminators alongside generators featuring 4, 6, and 9 residual blocks.

\vspace*{0.1cm}\noindent\textbf{Multi-scale perceptual discriminator:} It was implemented as a PatchGAN-based model and considers representations extracted at different output scales (multi-scale). Concretely, our discriminator classifies their inputs using overlapping square patches of 16, 8 and 4 sizes. This makes the learning process focus from the lowest-level to the higher-level image cues. Different from~\cite{Sungatullina2018}, we rely on VGG-16~\cite{Simonyan2015} to extract activations at four different scales.

\subsection{Loss functions}
Let $R$ and $S$ be the real and synthetic domains of hand images where $r_i\in R$ and $s_i\in S$ are samples from each domain. We denote the data distribution as $\boldsymbol{r} \sim p_{\mathrm{data}}(\boldsymbol{r})$ and $\boldsymbol{s} \sim p_{\mathrm{data}}(\boldsymbol{s})$. On the one side, the goal is make generators $G_s$ and $G_r$ learn two mappings $G:R\to S$ and $F:S\to R$, respectively. On the other side, discriminator $D_r$ discriminates between image $r_i$ and translated image $F(s_i)$; likewise, discriminator $D_s$ aims to distinguish between $s_i$ and $G(r_i)$. Generators and discriminators are trained simultaneously, in an adversarial fashion. 

Following the recommendations in the original \gls{cyclegan}, we have used the least-squares loss\cite{Mao2017} to stabilize training. For instance, generator $G_s$ and discriminator $D_s$ are trained as follows:
\begin{equation}\begin{aligned}
\mathcal{L}_{\mathrm{GAN}}\left(D_{S}, R, S\right) &=\frac{1}{2} \mathbb{E}_{\boldsymbol{s} \sim p_{\text{data}}(\boldsymbol{s})}\left[(D_{S}(\boldsymbol{s})-1)^{2}\right] \\
&+\frac{1}{2} \mathbb{E}_{\boldsymbol{r} \sim p_{\boldsymbol{\text{data}}}(\boldsymbol{r})}\left[(D_{S}(G_{S}(\boldsymbol{r})))^{2}\right] \\
\mathcal{L}_{\mathrm{GAN}}\left(G_{S}, R, S\right)  &= \frac{1}{2}\mathbb{E}_{\boldsymbol{s} \sim p_{\boldsymbol{s}}(\boldsymbol{s})}\left[(D_{S}(G_{S}(\boldsymbol{r}))-1)^{2}\right]
\end{aligned}\end{equation}

In the case of the multi-scale perceptual discriminators, their inputs are features extracted from the VGG network at four different scales. Moreover, the outputs of the discriminator are also at multiple scales. Formally, $D_{S}(s)$ and $D_{S}(G_{S}(r))$ output vectors $Y$ and $Z$ of $n$ elements where $n$ is the total number of output scales. Under this configuration, $\mathcal{L}_{\mathrm{GAN}}\left(D_{S}, R, S\right)$ loss is computed as $\sum_{i=1}^{n}\frac{1}{2}\left(Y_{i}-1\right)^{2} + \frac{1}{2}\left(Z_{i}\right)^{2}$. Regarding cycle consistency and identity losses, we have used the implemented in the original \gls{cyclegan}~\cite{Zhu2017}.

\section{Results}
\begin{figure*}[!htb]
\centering
\includegraphics[width=0.95\textwidth]{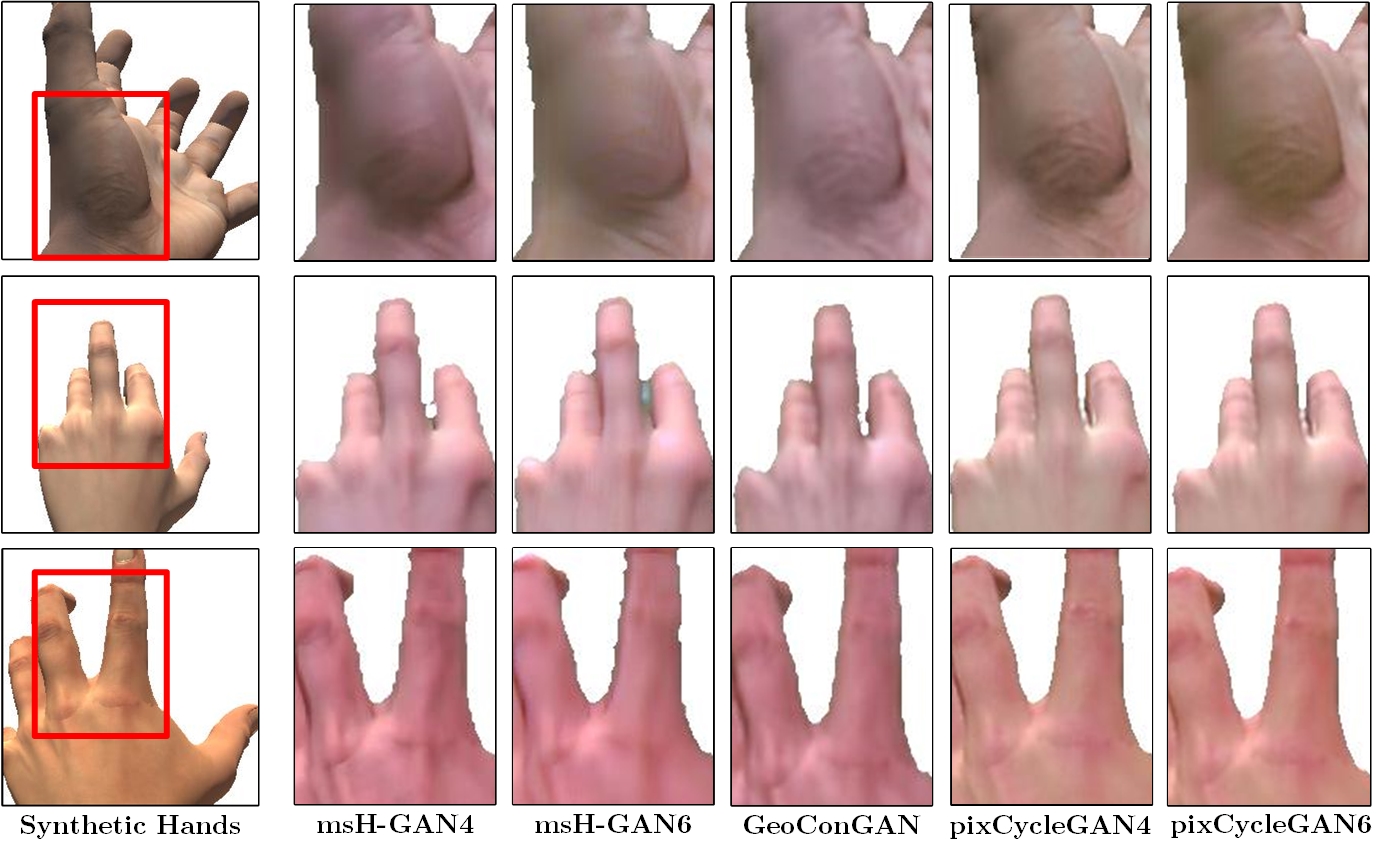}
\caption{Generated images with our proposed architectures and the \gls{geocongan} baseline. Notice how different are the outputs in terms of sharpness (top row), hand shape preservation (middle row) and skin color consistency (bottom row).}
\label{fig:hand_details}
\end{figure*}

In this section we define the model selection methodology and the evaluation procedure. Once selected, the best performing model configurations will be analyzed on a qualitative and quantitative basis. The qualitative evaluation consists of a visual inspection of the generated images in terms of skin color mapping realism, high-level details transfer and artifact presence. Furthermore, results will be compared with the baseline method using perceptual metrics.

\subsection{Implemented models and training details}
At the architectural level we have tested different generator/discriminator combinations. We varied the number of residual blocks in the generators and the discriminator type (patch, pixel and perceptual). Regarding the multi-scale perceptual discriminators we use outputs at different scales, concretely, squared patches of 16, 8 and 4 size. The implemented models are the following:
\begin{itemize}
    \item \textbf{\gls{geocongan}}~\cite{Mueller2018}. Baseline model with a patch discriminator and a ResNet generator with 9 residual blocks.
    \item \textbf{pixCycleGAN4}. Implements a ResNet generator with 4 residual blocks and a pixel discriminator.
    \item \textbf{pixCycleGAN6}. Different from pixCycleGAN4 it uses 6 residual blocks in the generator.
    \item \textbf{msH-GAN4}. Features a multi-scale perceptual discriminator and a ResNet with 4 residual blocks.
    \item \textbf{msH-GAN6}. Different from msH-GAN4, it uses a 6 residual blocks in the generator.
\end{itemize}

In order to select the best performing models, an exhaustive testing with different hyperparameter configurations was carried out. We explored changes in: learning rates, the identity loss scale factor to seek for the best diversity/fidelity ratio in the generated data, image sizes for training and validation, and training batch size, among others.

\vspace*{0.1cm}\noindent\textbf{Baseline:} \gls{geocongan} model is the latest work developed for the task at hand. Therefore, it is the baseline for our \gls{hgan} model for comparison purposes. \gls{geocongan} was trained following the original implementation details~\cite{Mueller2018}. For the unspecified hyperparameters, we have used the default \gls{cyclegan} configuration. Concretely, it is based on a Resnet generator with 9 residual blocks and a patch discriminator, both trained with a learning rate of $0.0002$. As a difference, our models do not rely on their hand shape-based loss. We directly use the provided masks in the dataset (with a previous pre-processing to remove noise and artifacts) to mask the generated images. 

\vspace*{0.1cm}\noindent\textbf{Training details:} All the models were trained with random cropped input images of $128\times128$ size and validated on bigger images of $256\times256$ using the \gls{fid} metric. Discriminators and generators were trained with a $0.00005$ learning rate on the whole datasets for 10 epochs. Networks were initialized using Xavier. Regarding weighted losses, we have used $0.5$ and $10$ weights for the identity loss and cycle loss, respectively. No learning rate schedulers were used. All the models were implemented using PyTorch Lightning framework and trained on NVIDIA TITAN Xp/V GPUs.

\subsection{Evaluation procedure}
We conducted an in-depth model search procedure to identify the best hyperparameter configurations. As a first step, a visual inspection of the results was carried out to broadly prune the less promising configurations. The fine-grain selection was performed validating the models using the \gls{fid} metric. Once training is completed, a quantitative evaluation is performed on the test set (2000 images) using the \gls{fid} and \gls{kid} metrics.

\vspace*{0.1cm}\noindent\textbf{Metrics:} Performance evaluation of generative models is challenging~\cite{Shmelkov2018,Xu2018} because of the lack of an explicit likelihood measure. Visual inspection to evaluate generated data quality is still a de facto evaluation protocol in the domain adaptation literature. However, some perceptual metrics such as \gls{fid}~\cite{Heusel2017} and \gls{kid}~\cite{Binkowski2018} were used as ad-hoc metrics correlated with the quality of the generated images. Particularly, \gls{fid} compares Gaussian statistics computed from activations (extracted from Inception V3 network) between the generated images and a bunch of samples from the target distribution. Conversely to \gls{fid}, \gls{kid} serves as an unbiased evaluation metric. Since both metrics are mere statistical approximations to the true difference, they guarantee no correlation with performance on real-world tasks~\cite{Shmelkov2018}. However, they empirically demonstrated their usefulness in quantitative evaluations. Therefore, our quantitative analysis stands on these two metrics.

\vspace*{0.1cm}\noindent\textbf{Datasets}: For training and validation we have used the SynthHands~\cite{Mueller2017} and RealHands~\cite{Mueller2018} datasets. The former provides synthetically generated images from an egocentric view. It contains fully-annotated images of male and female hands, both with and without object interaction. However, hand textures show undesirable artifacts and uncanny lighting or skin color. From the real domain, the RealHands dataset provides hand images of 7 different subjects with different skin tones and hand shapes. They were captured using a low-resolution webcam on a green-screen setup. Under this configuration, some images and their segmentation masks are blurry and contain strange artifacts. To use the masks in our models, we have refined them as follows. First, we reduced speckle noise using median filtering. Next, we addressed artifacts by searching for contours in the masks and removing those of a smaller area assuming the hand is the biggest contour. Finally, we performed morphological operations such as closing, to fill some gaps in the hand mask, and erosion to remove residual background pixels on the hand border.

\subsection{Qualitative and quantitative analysis}
\begin{table}[!tb]
    \centering
	\ra{1.0}
	\caption{Perceptual metrics evaluating the implemented models. The msH-GAN4 and msH-GAN6 models obtained better results compared to the \gls{geocongan} baseline and vanilla \glspl{cyclegan}.}
	\label{table:models}
	\resizebox{0.8\linewidth}{!}{
		\begin{tabular} {@{}lcc@{}} 
			\toprule
			\textbf{method} & \textbf{FID}$\downarrow$ & \textbf{KID}$\downarrow$ \\
			\midrule
			pix\gls{cyclegan}6 & $71.639$ & $0.0530\pm0.0065$ \\
			pix\gls{cyclegan}4 & $70.815$ & $0.0517\pm0.0065$ \\
			\gls{geocongan} & $66.786$ & $0.0481\pm0.0065$ \\
			ms\gls{hgan}4 & $63.297$ & $\mathbf{0.0479\pm0.0067}$ \\
			ms\gls{hgan}6 & $\mathbf{63.185}$ & $0.0480\pm0.0066$ \\
			\bottomrule
	\end{tabular}}
\end{table}

The qualitative and quantitative evaluation was performed on the test set containing 2000 images of hands. First, a rough visual inspection was performed to determine the overall consistency in the skin color mapping over the whole test set. In this regard, perceptual-based approaches have shown greater consistency. This means that the generated hands are in general of a higher visual quality showing more similarities in the skin color with the real images of hands. On the other hand, approaches using pixel discriminators (pixCycleGAN4 and pixCycleGAN6) reported sharper hands (see top row in Figure~\ref{fig:hand_details}) and a more consistent hand shape preservation in the translation. The latter was detected by inspecting the small gaps between hand fingers (see middle row in Figure~\ref{fig:hand_details}). In this regard, pixel discriminators are helpful since they make the decision at the pixel-level. However, they also tend to one-to-one map the synthetic skin tone and geometry artifacts to the real domain. As we can observe in Figure~\ref{fig:hand_details}, the differences are noteworthy. This difference is also reflected in both \gls{fid} and \gls{kid} metrics, reporting worse quantitative results.

On the other side, PatchGAN-based discriminators tend to fill the tiny gaps between hand fingers (see msH-GAN6, middle row sample in Figure~\ref{fig:hand_details}). To combat this side-effect, the GeoConGAN architecture relies on a UNet network (called SilNet) to segment the hand shape which is then used in their geometrically-consistent loss function. However, the differences with the msH-GAN4 model in terms of hand shape preservation are not clear. The latter uses the preprocessed hand masks which are applied to the generated images instead of an external architecture. This makes the model end-to-end and more parameter efficient avoiding the extra memory overhead.

Quantitatively, perceptual-based approaches reported better results than the \gls{geocongan}~\cite{Mueller2018} model. However, very subtle differences were noticed on the \gls{kid} metric. Furthermore, we claim that, for the problem at hand, the difference in the number of residual blocks in the generators has no major relevance to performance. Overall, the best performing methods were the msH-GAN4 and msH-GAN6 models. However, msH-GAN4 presented sharper results, subtle differences in skin color (visually more similar to the real hands) and was also less prominent in filling the gaps between fingers.

\subsection{A note on cross-domain similarity}
\label{subsec:applications}
\begin{figure}[!t]
\centering
\includegraphics[width=\linewidth]{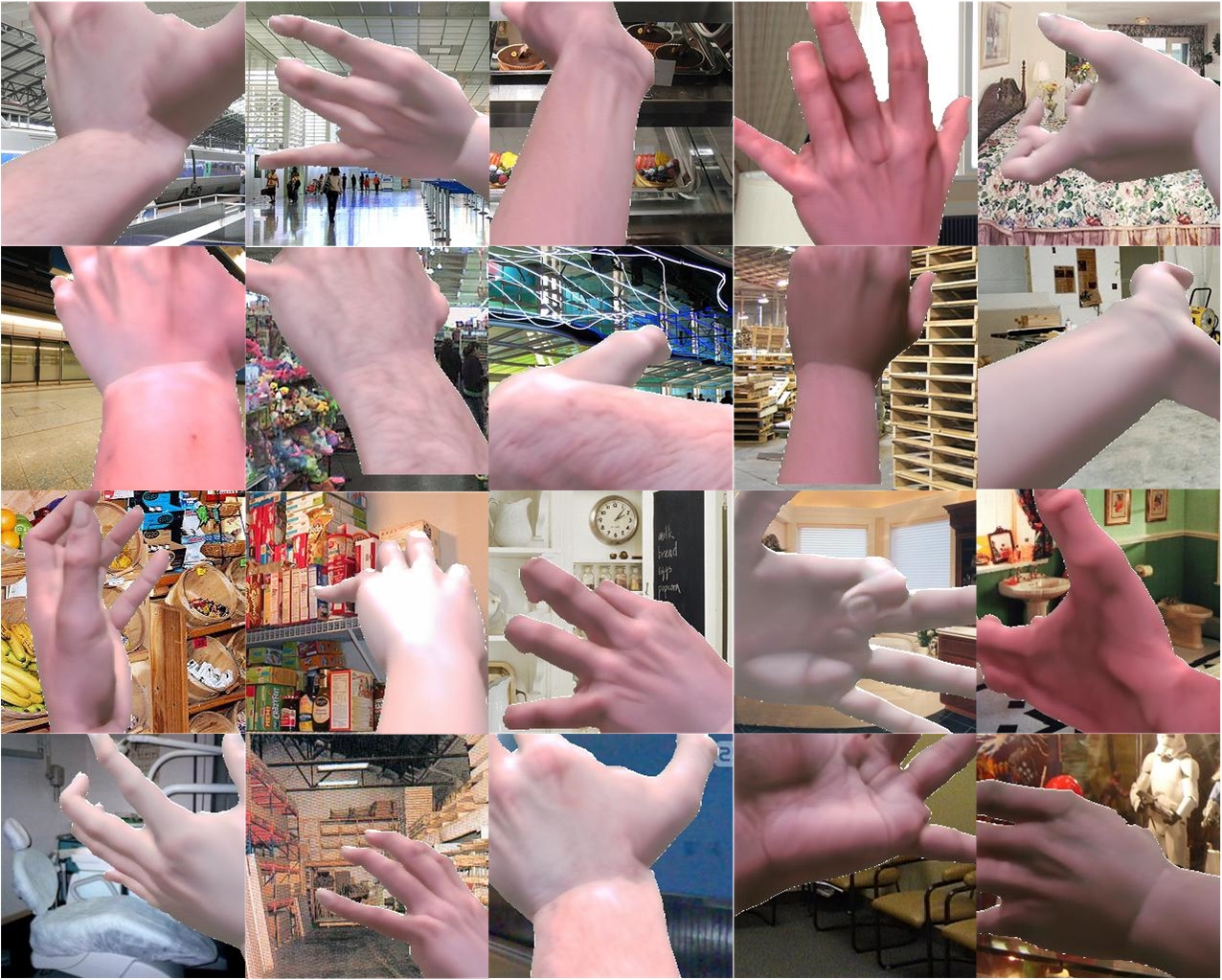}
\caption{Random subset of the generated hands used in the classification task. Hands were generated using our \mbox{msH-GAN6} model and placed on random backgrounds extracted from~\cite{Quattoni2009}.}
\label{fig:hand_application}
\end{figure}
\begin{table*}[!htb]
    \centering
	\ra{1.1}
	\caption{Evaluation of the \acrshort{vit} model trained for binary classification (synthetic or real) and tested with synthetic (syn), real (rea) and generated (gen) hands. For training, we assume real and generated hands belong to the same class (real). Notice both classes were classified indistinguishably reporting slight differences in the metrics. $\|$ denotes separation between classes.}
	\label{table:test1}
	\resizebox{0.8\linewidth}{!}{
		\begin{tabular} {@{}lllcccc@{}} 
			\toprule
			\textbf{$\#$} & \textbf{train set} & \textbf{test set} & \textbf{precision}$\uparrow$ & \textbf{recall}$\uparrow$ & \textbf{f1-score}$\uparrow$ & \textbf{accuracy}$\uparrow$ \\
			\midrule
			1 & syn $\|$ rea & syn $\|$ rea & $94.51\,\|\,97.38$ & $97.46\,\|\,94.34$ & $95.96\,\|\,95.84$ & $95.90$ \\
			& & syn $\|$ gen & $95.62\,\|\,97.18$ & $97.23\,\|\,95.54$ & $96.42\,\|\,96.35$ & $96.39$ \\
			\midrule
			2 & syn $\|$ gen & syn $\|$ rea & $95.29\,\|\,92.65$ & $92.43\,\|\,95.43$ & $93.84\,\|\,94.02$ & $93.93$ \\
			& & syn $\|$ gen & $98.91\,\|\,93.60$ & $93.23\,\|\,98.97$ & $95.98\,\|\,96.21$ & $96.10$\\
			\midrule
			3 & syn $\|$ (\textonehalf\,rea, \textonehalf\,gen) & syn $\|$ rea & $94.83\,\|\,96.96$ & $97.03\,\|\,94.71$ & $95.92\,\|\,95.82$ & $95.87$ \\
			& & syn $\|$ gen & $97.25\,\|\,97.15$ & $97.14\,\|\,97.26$ & $97.20\,\|\,97.20$ & $97.20$\\
			\bottomrule
	\end{tabular}}
\end{table*}

We rely on the hand classification task to claim our generated hands are statistically similar to the real domain of hands. We assume generated and real hands approximate a similar distribution if a classifier: (a) without being trained on generated hands, classifies most of them as real hands, and (b) confuses real with generated hands and vice versa, when learning to discriminate between them. We use metrics such as precision, recall, f1-score and accuracy, along with a confusion matrix showing the number of misclassified images, to analyze how similar is our generated data to the real domain of hands.

For (a), we have trained the \acrfull{vit}~\cite{Dosovitskiy2020} model to distinguish between synthetic and real images of hands. Without the model knowing what a generated hand looks like, the goal is to classify the generated images and determine to which class (synthetic or real) they belong. We assume that if most of the generated images are classified as real hands, they approximate a similar distribution. Three different training sets with equal contribution per class were used in this first experiment:
\begin{enumerate}
    \item $10.500$ synthetic and $10.500$ real images.
    \item $10.500$ synthetic and $10.500$ generated images.
    \item $10.500$ synthetic and a combination of $5.250$ real and $5.250$ generated images representing the real hand class.
\end{enumerate}
The real and generated images were classified as real hands with similar precision in all tested scenarios (see Table~\ref{table:test1}). Furthermore, the use of a combination of generated and real images in the training set showed better results in classifying the generated images, while maintaining similar performance in classifying real hands.
\begin{table}[!tb]
    \centering
	\ra{1.1}
	\caption{Evaluation of the \acrshort{vit} model trained on synthetic, real and generated images of hands ($10.500$ images per class). Notice how the model precisely classifies the synthetic hands, while confuses the real and generated classes reporting lower precision, recall and f$1$-score.}
	\label{table:test2}
	\resizebox{0.75\linewidth}{!}{
		\begin{tabular} {@{}lccc@{}} 
			\toprule
			\textbf{class} & \textbf{precision}$\uparrow$ & \textbf{recall}$\uparrow$ & \textbf{f1-score}$\uparrow$\\
			\midrule
			synthetic & $0.9540$ & $0.9546$ & $0.9543$ \\ 
			real & $0.7930$ & $0.8329$ & $0.8124$ \\
			generated & $0.8468$ & $0.8037$ & $0.8247$ \\
			\midrule
			\midrule
			\textbf{accuracy} & & & 0.8637 \\
			\bottomrule
	\end{tabular}}
\end{table}

But, what if we consider the generated hands belong to a different class? For (b) we want the \gls{vit} model to learn the differences between real and generated hands. Note that the generated and synthetic hands have the same shape, but a different appearance. This shape correspondence overcomplicates the classification task and makes the \gls{vit} model focus on the appearance of the hand rather than its shape. Assuming that the generated and real hands share similarities at a distribution level, the model should confuse these two classes, i.e. misclassify a generated hand as real and vice versa. We show the number of misclassified samples in the confusion matrix (see Figure~\ref{fig:confusion}) and the model performance in terms of precision, recall and f$1$-score in Table~\ref{table:test2}. The \gls{vit} model precisely ($95.40\%$) classifies the synthetic hands, yet it confuses the generated with the real hands and vice versa. Concretely, $647$ ($18\%$) of the real hands were classified as generated and $465$ ($13\%$) of the generated hands as real. The misclassification has a direct impact on the per-class precision, recall and f$1$-score revealing the similarities between the generated and real hands.

Models were trained for $20$ epochs with a learning rate of $0.00001$ and tested with $3.500$ samples per class. For more implementation details, see the code repository.

\begin{figure}[!t]
\centering
\resizebox{.75\linewidth}{!}{\def\myConfMat{{
{3341,  121,   40},  %row 1
{ 114, 2915,  647},  %row 2
{  45,  465, 2813},  %row 3
}}

\def\classNames{{"syn","rea","gen"}} %class names. Adapt at will

\def\numClasses{3} %number of classes. Could be automatic, but you can change it for tests.

\def\myScale{1.5} % 1.5 is a good scale. Values under 1 may need smaller fonts!
\begin{tikzpicture}[
    scale = \myScale,
    %font={\scriptsize}, %for smaller scales, even \tiny may be useful
    ]

\tikzset{vertical label/.style={rotate=90,anchor=east}}   % usable styles for below
\tikzset{diagonal label/.style={rotate=45,anchor=north east}}

\foreach \y in {1,...,\numClasses} %loop vertical starting on top
{
    % Add class name on the left
    \node [anchor=east] at (0.4,-\y) {\pgfmathparse{\classNames[\y-1]}\pgfmathresult}; 
    
    \foreach \x in {1,...,\numClasses}  %loop horizontal starting on left
    {
%---- Start of automatic calculation of totSamples for the column ------------   
    \def\totSamples{0}
    \foreach \ll in {1,...,\numClasses}
    {
        \pgfmathparse{\myConfMat[\ll-1][\x-1]}   %fetch next element
        \xdef\totSamples{\totSamples+\pgfmathresult} %accumulate it with previous sum
        %must use \xdef fro global effect otherwise lost in foreach loop!
    }
    \pgfmathparse{\totSamples} \xdef\totSamples{\pgfmathresult}  % put the final sum in variable
%---- End of automatic calculation of totSamples ----------------
    
    \begin{scope}[shift={(\x,-\y)}]
        \def\mVal{\myConfMat[\y-1][\x-1]} % The value at index y,x (-1 because of zero indexing)
        \pgfmathtruncatemacro{\r}{\mVal}   %
        \pgfmathtruncatemacro{\p}{round(\r/\totSamples*100)}
        \coordinate (C) at (0,0);
        \ifthenelse{\p<50}{\def\txtcol{black}}{\def\txtcol{white}} %decide text color for contrast
        \node[
            draw,                 %draw lines
            text=\txtcol,         %text color (automatic for better contrast)
            align=center,         %align text inside cells (also for wrapping)
            fill=black!\p,        %intensity of fill (can change base color)
            minimum size=\myScale*10mm,    %cell size to fit the scale and integer dimensions (in cm)
            inner sep=0,          %remove all inner gaps to save space in small scales
            ] (C) {\r\\\p\%};     %text to put in cell (adapt at will)
        %Now if last vertical class add its label at the bottom
        \ifthenelse{\y=\numClasses}{
        \node [] at ($(C)-(0,0.75)$) % can use vertical or diagonal label as option
        {\pgfmathparse{\classNames[\x-1]}\pgfmathresult};}{}
    \end{scope}
    }
}
%Now add x and y labels on suitable coordinates
\coordinate (yaxis) at (-0.4,0.4-\numClasses/2);  %must adapt if class labels are wider!
\coordinate (xaxis) at (0.5+\numClasses/2, -\numClasses-1.15); %id. for non horizontal labels!
\node [vertical label] at (yaxis) {Predicted Class};
\node []               at (xaxis) {True Class};
\end{tikzpicture}}
\caption{Confusion matrix representing the number of mislabelled samples among the different classes: synthetic (syn), real (rea), and generated (gen). Notice the confusion between the real and generated classes. $18\%$ of real hands were classified as generated, and $13\%$ of generated hands as real.}
\label{fig:confusion}
\end{figure}

\section{Conclusions}
Despite the wealth of approaches in the domain adaptation field, methods addressing the synthetic-to-real translation of hands are few. On this line, we proposed the \gls{hgan} model aiming to provide the perfect blend of a realistic hand appearance with synthetic annotations. We showed our \gls{hgan} tackles not only inter-domain tone mapping but also hallucinates finer details. After an exhaustive model selection procedure, we reported comparable results to previous approaches on a qualitative and quantitative basis. We outperformed the \gls{geocongan}~\cite{Mueller2018} model on \gls{fid} and \gls{kid} metrics. Furthermore, the generated data was used in a classification network demonstrating its usefulness in real-world applications.

\section*{Acknowledgment}
This work has been funded by the Spanish Government PID2019-104818RB-I00 grant for the MoDeaAS project, supported with Feder funds. This work has also been supported by two Spanish national grants for PhD studies, FPU17/00166, and ACIF/2018/197 respectively. We would also like to thank Nvidia for their generous hardware donation that made these experiments possible.



%\bibliographystyle{IEEEtran}
\bibliographystyle{IEEEtran}
\bibliography{references}

\end{document}